\documentclass[10pt,twocolumn,letterpaper]{article}

\usepackage{cvpr}              

\newcommand{\TODO}[1]{\textbf{\color{red}[TODO: #1]}}
\renewcommand{\TODO}[1]{}

\usepackage{microtype}

\usepackage{tikz}
\usepackage{multirow}
\usepackage{comment}
\usepackage{booktabs}
\usepackage{amssymb}
\usepackage{amsmath}
\usepackage{graphicx}
\usepackage{pifont}
\usepackage[x11names]{xcolor}

\definecolor{cvprblue}{rgb}{0.21,0.49,0.74}
\usepackage[pagebackref,breaklinks,colorlinks,allcolors=cvprblue]{hyperref}

\def\paperID{*****} 
\def\confName{CVPR}
\def\confYear{2026}

\title{OmniLight: One Model to Rule All Lighting Conditions}

\author{Youngjin Oh$^{1}$ \quad Junyoung Park$^{1}$ \quad Junhyeong Kwon$^{1}$ \quad Nam Ik Cho$^{1,2}$\\
$^{1}$Department of ECE, INMC, Seoul National University, South Korea\\
$^{2}$IPAI, Seoul National University, South Korea\\
{\tt\small \{yjymoh0211, parkjun210, gjh8760, nicho\}@snu.ac.kr}
}

\begin{document}

\twocolumn[{
\maketitle
\vspace{-8mm}
\begin{center}
    \captionsetup{type=figure}
    \setlength{\abovecaptionskip}{2mm}
    \centering
    \includegraphics[width=0.98\textwidth]{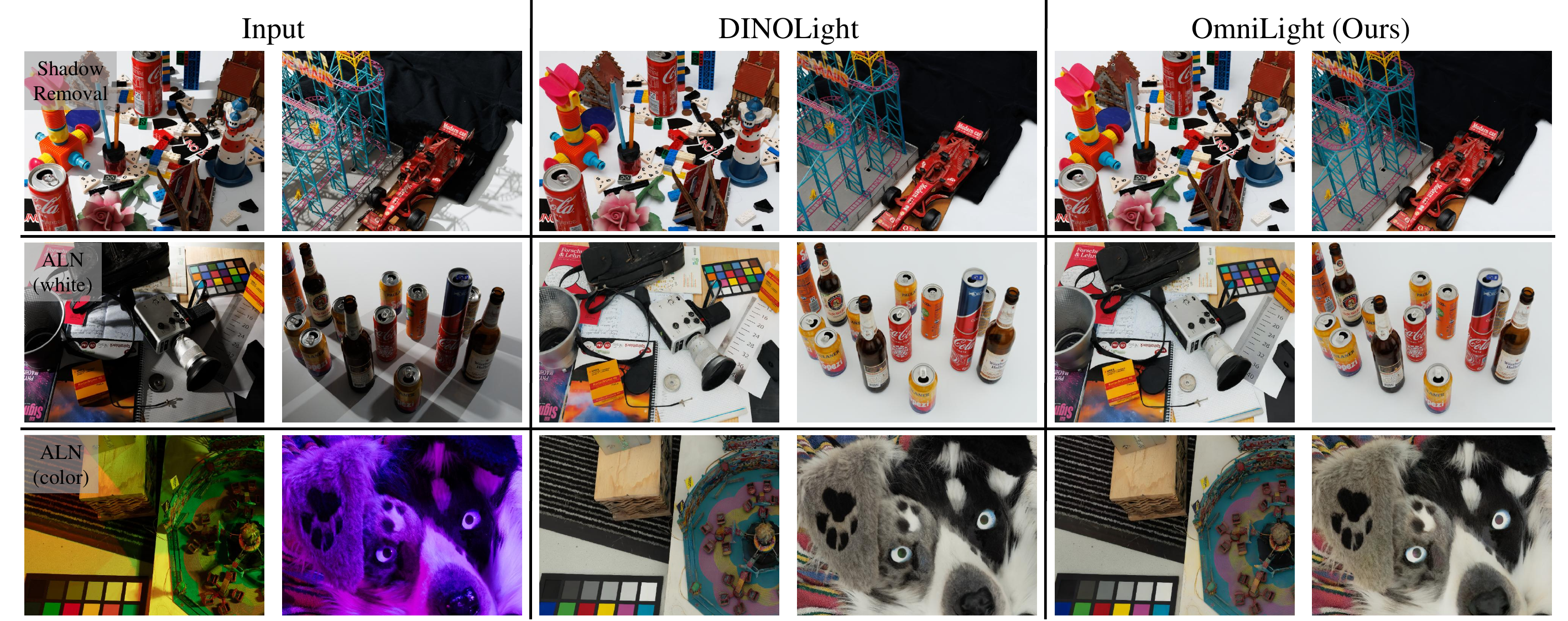}
    \caption{Restoration examples of the baseline DINOLight~\cite{oh2026dinolight} and the proposed unified method OmniLight. In the NTIRE 2026 Challenges~\cite{ntire2026ambient,ntire2026shadow}, OmniLight ranked \textbf{\textcolor{red}{1st (Perceptual)}} in ALN White Lighting, and \textbf{\textcolor{blue}{2nd (Fidelity)}} and \underline{4th (Perceptual)} in ALN Color Lighting. Furthermore, DINOLight ranked \textbf{\textcolor{red}{1st (Fidelity)}} and \textbf{\textcolor{blue}{2nd (Perceptual)}} in ALN Color Lighting, and \textbf{3rd} in Shadow Removal.}
    \label{figure:figure_firstpage}
\end{center}
}]

\begin{abstract}
Adverse lighting conditions, such as cast shadows and irregular illumination, pose significant challenges to computer vision systems by degrading visibility and color fidelity. Consequently, effective shadow removal and ALN are critical for restoring underlying image content, improving perceptual quality, and facilitating robust performance in downstream tasks. However, while achieving state-of-the-art results on specific benchmarks is a primary goal in image restoration challenges, real-world applications often demand robust models capable of handling diverse domains. To address this, we present a comprehensive study on lighting-related image restoration by exploring two contrasting strategies. We leverage a robust framework for ALN, DINOLight, as a specialized baseline to exploit the characteristics of each individual dataset, and extend it to OmniLight, a generalized alternative incorporating our proposed Wavelet Domain Mixture-of-Experts (WD-MoE) that is trained across all provided datasets. Through a comparative analysis of these two methods, we discuss the impact of data distribution on the performance of specialized and unified architectures in lighting-related image restoration. Notably, both approaches secured top-tier rankings across all three lighting-related tracks in the NTIRE 2026 Challenge, demonstrating their outstanding perceptual quality and generalization capabilities. Our codes are available at \url{https://github.com/OBAKSA/Lighting-Restoration}.
\end{abstract}
    
\section{Introduction}
\label{sec:intro}

The restoration of images degraded by complex, non-uniform lighting conditions constitutes one of the most persistent and ill-posed challenges in low-level computer vision. While the field of image restoration has witnessed substantial maturity in sub-domains such as denoising~\cite{abdelhamed2018high,plotz2017benchmarking,kim2020transfer,zamir2022restormer,chen2022simple,kim2025idf}, super-resolution~\cite{soh2020meta,liang2021swinir,conde2022swin2sr,chen2021learning,yu2024scaling,chu2022nafssr}, and single-image shadow removal~\cite{vasluianu2023wsrd,le2019shadow,wang2018recovering,qu2017deshadownet,le2020shadow,cun2020towards,wang2018stacked,jin2021dc,jin2024des3,zhang2022spa,guo2023shadowformer,dong2024shadowrefiner,xiao2024homoformer,hu2019mask,vasluianu2021shadow,liu2021shadow,guo2023shadowdiffusion,mei2024latent,guo2023boundary,xu2025omnisr,vasluianu2024ntire}, the broader and more physically intricate problem of Ambient Lighting Normalization (ALN)~\cite{vasluianu2024towards,vasluianu2025after} remains an open frontier. Unlike conventional shadow removal, which typically assumes a binary occlusion model under a single dominant light source, ALN necessitates the restoration of images suffering from multiple interacting illuminants, complex scene geometries inducing self-shadowing, and diverse surface material reflectance. The recent emergence of specialized benchmark datasets, specifically Ambient6K~\cite{vasluianu2024towards} and CL3AN~\cite{vasluianu2025after}, has fundamentally shifted the research paradigm. These benchmarks demand models capable of disentangling intrinsic scene properties (reflectance) from extrinsic illumination factors (shading, chromatic cast) with unprecedented fidelity, moving beyond simple binary mask-guided restoration to comprehensive lighting normalization.

Motivated by the NTIRE workshop challenge tracks \textit{Shadow Removal, ALN (White Lighting)}, and \textit{ALN (Color Lighting)}~\cite{ntire2026ambient,ntire2026shadow}, this paper presents a comprehensive study on all-in-one restoration for complex lighting conditions. In this work, we formally categorize shadow removal, white-balanced ALN, and multi-color ALN under a unified framework, defining them collectively as \textit{lighting-related image restoration}. We investigate two distinct methodological approaches: Specialized Task Learning, represented by an existing robust baseline framework \textbf{DINOLight}~\cite{oh2026dinolight}, and Unified Learning, embodied by the proposed \textbf{OmniLight} architecture. To our knowledge, this is the first work to tackle lighting-related image restoration with a holistic all-in-one approach.

DINOLight~\cite{oh2026dinolight} offers a highly tailored solution for lighting-related restoration tasks. It operates on the hypothesis that the ambiguity of lighting in complex scenes, where a dark region could be a shadow, a dark object, or a void, can be effectively resolved through the injection of potent visual priors. By integrating features from large-scale self-supervised foundation models, specifically DINOv2~\cite{oquab2023dinov2}, DINOLight excels in deep shadows and resolving geometric ambiguities within the specific distribution of the Ambient6K dataset. However, its heavy specialization on a single training dataset potentially limits its adaptability to out-of-distribution scenarios, such as the aggressive chromatic shifts introduced in multi-colored lighting cases.

Conversely, OmniLight takes a generalized approach, applying universal lighting principles across different datasets. To capture this universality while mitigating the ``negative transfer" phenomena that often hamper multi-task learning, OmniLight employs our proposed Wavelet Domain Mixture-of-Experts (WD-MoE) branch to aid the main DINOLight branch. This framework decomposes the image features into frequency sub-bands, separating low-frequency illumination trends from high-frequency textural details, and routes these components to specialized expert sub-networks. The low- and high-band experts refine the features guided by the degradation-aware routing vectors, and the features are combined to modulate the DINOLight branch.

Our main contributions are as follows:
\begin{itemize}
    \item We formally unify the isolated tasks of single-image shadow removal, white-balanced ALN, and multi-color ALN under a single problem domain. To the best of our knowledge, this work introduces the first holistic, all-in-one approach designed to comprehensively tackle the wide spectrum of complex illumination degradations.
    \item We propose OmniLight, a unified network that mitigates ``negative transfer" across diverse tasks. By utilizing our Wavelet Domain Mixture-of-Experts (WD-MoE) with the degradation-aware routing guidance vector, the frequency-decoupled features are effectively refined to modulate the main branch, achieving universal lighting-related image restoration.
    \item We empirically validate the robustness and performance of our unified approach OmniLight in the NTIRE 2026 challenge~\cite{ntire2026ambient,ntire2026shadow}. Our proposed OmniLight, based on a robust ALN solution~\cite{oh2026dinolight}, secured top performance in the challenges, demonstrating its strong generalization capabilities to lighting-related image restoration tasks.
\end{itemize}

\section{Related Works}
\label{sec:related}

\subsection{Datasets for Lighting-Related Image Restoration: From Shadow Removal to ALN}
Early deep learning-based research in this domain was limited by data and computation, focusing mainly on simple shadow removal. Datasets like ISTD~\cite{wang2018stacked}, ISTD+~\cite{le2019shadow} and SRD~\cite{qu2017deshadownet} assumed basic conditions: a single light source, flat surfaces, and uniform lighting. Most methods that show great performance on these datasets~\cite{le2019shadow,wang2018stacked,cun2020towards,vasluianu2021shadow,vasluianu2023wsrd,fu2021auto,guo2023shadowformer,guo2023shadowdiffusion,vasluianu2024ntire,hu2018direction,li2025shadowmaskformer,liu2024regional,wan2022style,xiao2024homoformer,zhu2022bijective} heavily rely on binary masks to separate ``shadow" and ``non-shadow" regions, reducing the task to simply matching colors between the two. Later on, a more developed dataset WSRD~\cite{vasluianu2023wsrd} was introduced with complex light-object interactions compared to previous shadow removal datasets. While effective for basic shadows, this approach fails in real-world ALN, where shadows are not simple binary defects but continuous transitions caused by complex 3D object shapes and lighting.

The Ambient6K dataset~\cite{vasluianu2024towards} significantly advanced the field by introducing the ALN task with high-resolution images under complex, multi-source lighting. This dataset demonstrated that mask-based methods are insufficient. In scenes with multiple lights and complex objects (e.g., crumpled fabric or houseplants), shadows vary continuously in intensity and softness depending on the object's shape. To succeed, models must understand the scene's 3D structure and its semantics to distinguish between an inherently dark surface (e.g., black paint) and an actual shadow. This need for both semantic and geometric understanding motivates the usage of visual priors, which is the core strategy of several proposed methods for ALN~\cite{vasluianu2025ntire}, such as DINOLight~\cite{oh2026dinolight} and PromptNorm~\cite{serrano2025promptnorm}. While some recent shadow removal methods like OmniSR~\cite{xu2025omnisr} also adopt vTisual priors, our approach distinguishes itself by explicitly leveraging the frequency domain and MoE for restoration.

Recently, the CL3AN dataset~\cite{vasluianu2025after} introduced the challenge of multi-color lighting, adding complex interactions between color and brightness. In Ambient6K, models could succeed by simply brightening dark areas or normalizing bright areas. However, in CL3AN, a surface might look red due to a red light source (illumination), not its actual color (reflectance). Therefore, a robust restoration model must independently adjust both color and brightness for each pixel (i.e., spatially varying chromatic adaptation). This represents a major shift from previous tasks; a model trained only on Ambient6K would mistake colored lighting for the object's true color, resulting in severe color distortion.

\subsection{All-in-One Image Restoration}
The field of image restoration has recently transitioned from task-specific networks~\cite{chen2022simple,chen2021hinet,zamir2022restormer,liang2021swinir,guo2024mambair,shi2025vmambair,guo2025mambairv2,cai2023retinexformer} toward all-in-one frameworks capable of addressing multiple degradation types, such as noise, haze, and rain, within a single, unified architecture~\cite{duan2024uniprocessor,zamfir2025complexity,conde2024instructir,potlapalli2023promptir,gao2024prompt,li2022all,park2023all,valanarasu2022transweather,zhang2023all}. A primary challenge in this unified paradigm is the efficient and adaptive allocation of model capacity across highly diverse restoration tasks. Notably, recent literature has explored dynamic routing mechanisms to address this constraint. For instance, Zamfir \textit{et al.}~\cite{zamfir2025complexity} proposed a Mixture-of-Complexity-Experts approach, utilizing flexible expert blocks with varying computational complexities to effectively perform task-discriminative routing for any image restoration.

However, while these architectures have demonstrated remarkable success, they have been predominantly designed for and evaluated on standard weather-induced degradations or spatially uniform corruptions (e.g., Gaussian noise, global low-light conditions). Complex, real-world illumination phenomena, characterized by multi-source color entanglement and geometry-dependent shadows, remain largely unexplored in this context. To the best of our knowledge, our work is the first to extend the generalized restoration paradigm to comprehensively tackle complex lighting-related tasks. By specifically addressing the spatially varying chromatic shifts and continuous penumbras inherent to lighting-related image restoration, we pioneer the integration of complex illumination correction into a unified restoration framework.
\section{Method}
\label{sec:method}

\begin{figure}[t]
\centering
   \includegraphics[width=7.5cm]{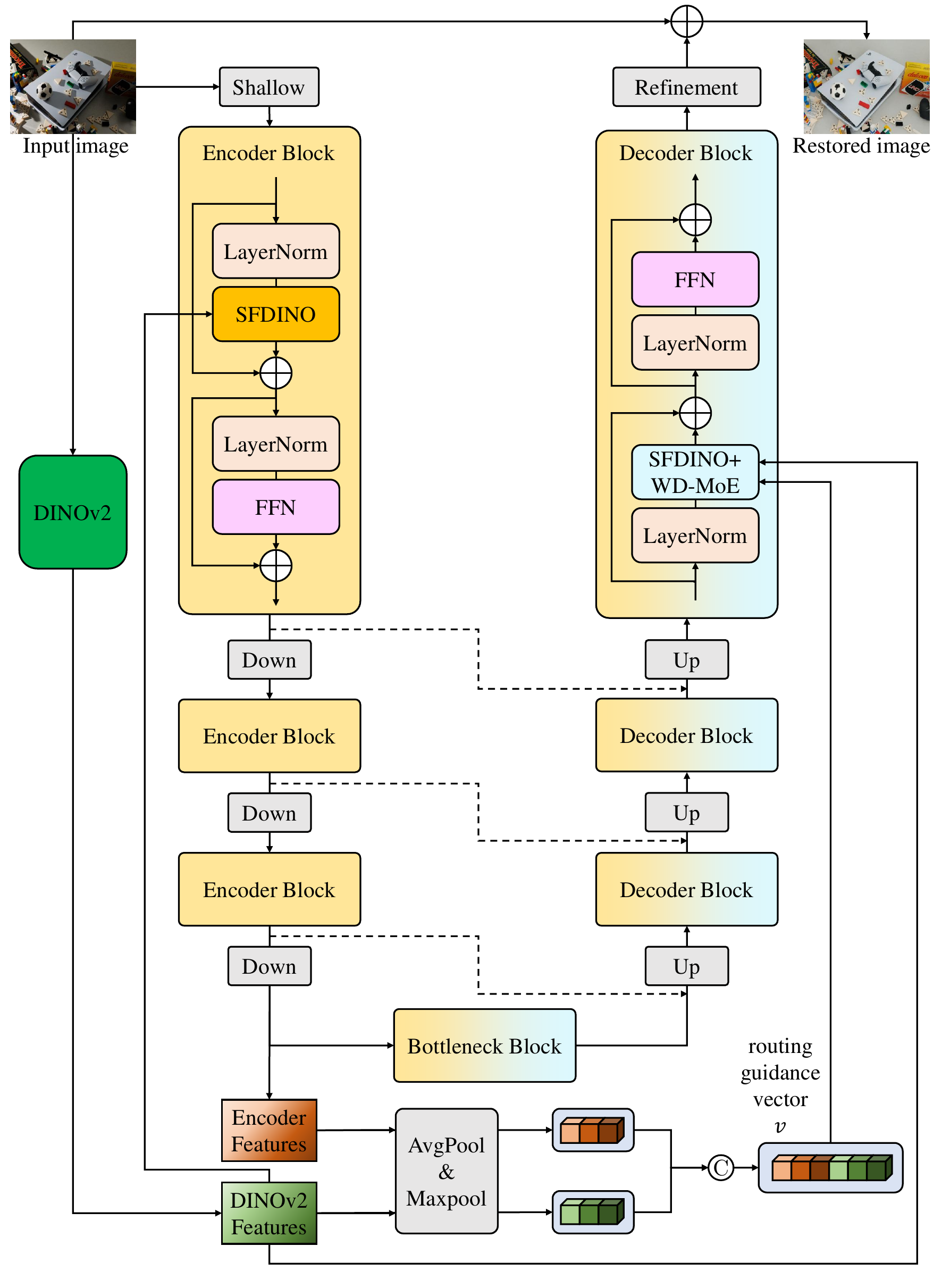}
   \hfil
\vspace{-3mm}
\caption{Overall architecture of the proposed OmniLight. The network employs a U-Net-based~\cite{ronneberger2015u} structure to process the input image, extracting multi-scale representations through a series of encoder and decoder blocks. While the encoder utilizes standard SFDINO blocks~\cite{oh2026dinolight}, the bottleneck and the decoder are additionally equipped with the proposed WD-MoE for degradation-aware feature conditioning. A degradation-aware routing guidance vector is explicitly computed with the concatenated encoder features and DINOv2 features for the router to utilize.}
\label{figure:figure_network}
\end{figure}

\subsection{Motivation and Overview}
While specialized architectures like the existing baseline DINOLight~\cite{oh2026dinolight} exhibit strong performance on specific illumination distributions by leveraging geometric and high-level semantic priors, they inevitably suffer from performance degradation when exposed to out-of-distribution complexities that were unseen during training. To transition from a task-specific model to a unified framework for all lighting-related image restoration tasks, it is crucial to minimize the ``negative transfer" phenomenon, where optimizing for one task leads to performance drop in another, for example, hard shadow boundaries (WSRD+) conflicting with spatially varying chromatic adaptation (CL3AN).

To this end, we propose OmniLight, a generalized All-in-One architecture. As illustrated in \cref{figure:figure_network}, the overall pipeline takes a degraded image as input, extracts robust geometric and semantic priors using a frozen DINOv2 encoder, and processes the features through a cascade of OmniLight Blocks. Instead of relying on a static, shared network, OmniLight adaptively conditions features based on degradation-aware input characteristics, thereby achieving universality across diverse lighting conditions.

\begin{figure}[t]
\centering
   \includegraphics[width=8cm]{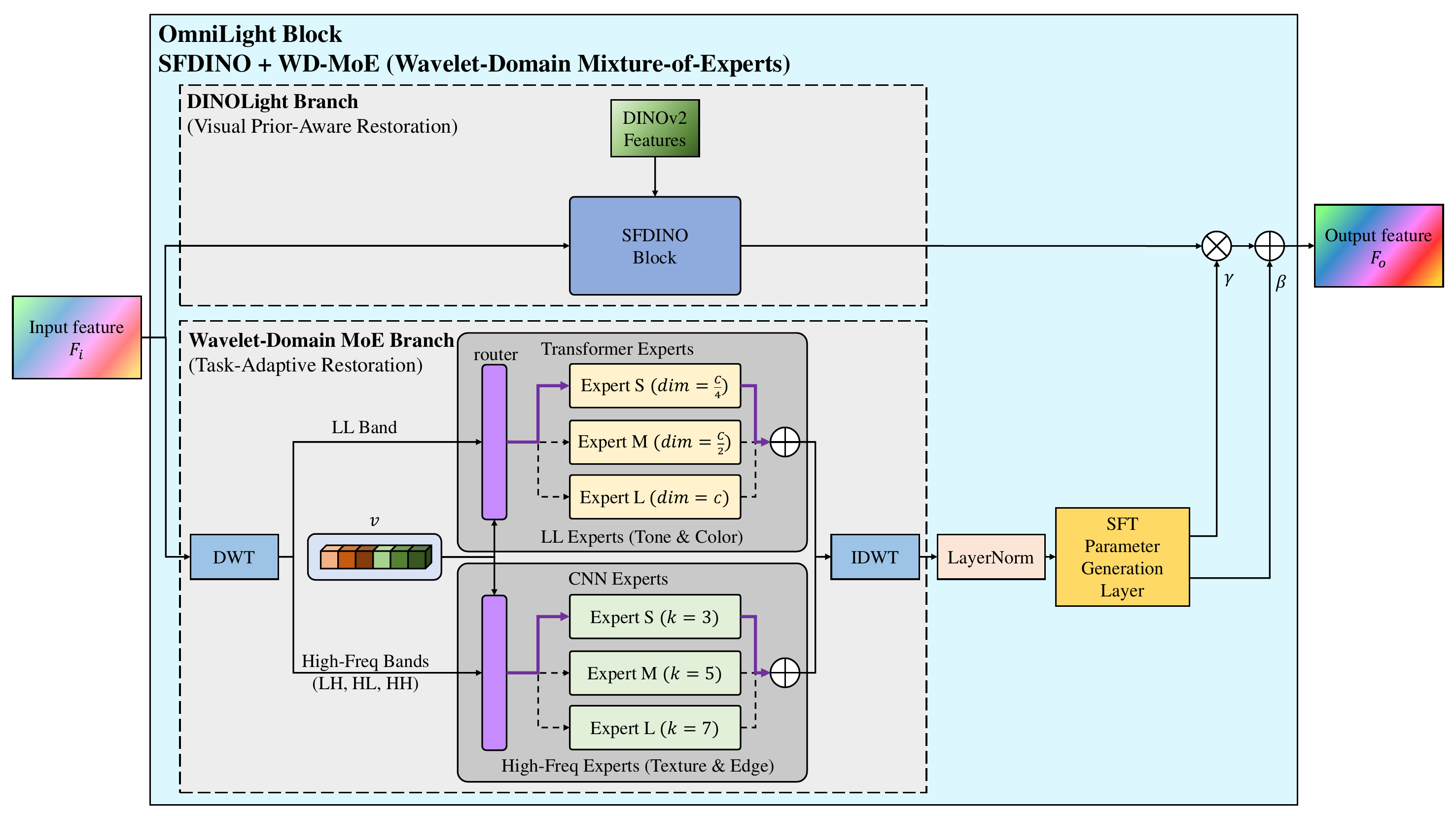}
   \hfil
\vspace{-2mm}
\caption{Detailed illustration of the dual-branched OmniLight block which consists of a DINOLight branch and a WD-MoE branch.}
\label{figure:figure_block}
\end{figure}

\subsection{OmniLight Block}
The core computational unit of our network is the OmniLight Block (depicted in \cref{figure:figure_block}), which adopts a dual-branch design to balance shared representation learning and degradation-aware refinement. The block consists of two primary components: the DINOLight Branch and the WD-MoE Branch.

The DINOLight branch serves as the backbone, continuously refining the spatial and semantic representations guided by the DINOv2 priors into the restoration process using SFDINO blocks~\cite{oh2026dinolight}. Concurrently, the WD-MoE branch operates as a conditioning module. It explicitly separates the incoming feature maps into distinct frequency sub-bands, processes them through expert sub-networks, and fuses the refined features back into the DINOLight branch. This dual-pathway allows the network to maintain a strong geometric and semantic understanding of the scene while adaptively refining the frequency features based on the specific requirements of each local patch.

\subsection{Wavelet Domain Mixture-of-Experts}
The lighting-related degradations exhibit distinct frequency behaviors: global color casts and soft penumbras predominantly reside in low-frequency bands, while sharp shadow edges and underlying textural details are high-frequency phenomena. 

To exploit this property, the WD-MoE branch first applies a 2D Discrete Wavelet Transform (DWT) to the input feature $X \in \mathbb{R}^{C \times H \times W}$, decomposing it into four sub-bands:
\begin{equation}
    X_{LL}, X_{LH}, X_{HL}, X_{HH} = \text{DWT}(X),
\end{equation}
where $X_{LL}$ represents the low-frequency approximation, and the rest represent high-frequency details, which we will refer to as $X_{H}=\text{Concat}(X_{LH}, X_{HL}, X_{HH})$. 

Rather than processing these components with a static convolution, we deploy a Mixture-of-Experts architecture. We define a set of low-frequency experts $E^{L} = \{E_{1}^{L}, \dots, E_{N}^{L}\}$ and high-frequency experts $E^{H} = \{E_{1}^{H}, \dots, E_{M}^{H}\}$. A gating router $G(\cdot)$ analyzes the global contextual embedding of the input to generate routing weights. The routed outputs for the low-band $\tilde{X}_{LL}$ and high-bands $\tilde{X}_{H}$ are computed as:
\begin{equation}
    \begin{aligned}
    \tilde{X}_{LL} = \sum_{i=1}^{N} G_{i}^{L}(X) \cdot E_{i}^{L}(X_{LL}), \\
    \tilde{X}_{H} = \sum_{j=1}^{M} G_{j}^{H}(X) \cdot E_{j}^{H}(X_{H}).
    \end{aligned}
\end{equation}

Following the expert routing, an Inverse DWT (IDWT) reconstructs the spatial feature map. Finally, this refined feature is used to modulate the output of the DINOLight branch via Spatial Feature Transform (SFT)~\cite{wang2018recovering}, effectively adjusting the robust backbone features with degradation-aware frequency refinements.

\noindent\textbf{Expert composition.}
Within the WD-MoE branch, the decomposed frequency sub-bands exhibit fundamentally different degradation characteristics. To effectively instantiate the expert sub-networks without introducing excessive computational overhead, we draw inspiration from established architectures in the image restoration literature~\cite{zamir2022restormer,chen2022simple}. We set the number of sub-networks for each branch $N$ and $M$ as 3, and allocate the complexities of each expert with varying numbers of channels or kernel sizes as shown in \cref{figure:figure_block}, motivated by MoCE-IR~\cite{zamfir2025complexity}.

For the low-frequency experts $E^{L}$, which are intended to handle tone, color, broad illumination trends and soft penumbras, we adopt Restormer~\cite{zamir2022restormer} blocks. As demonstrated in prior works~\cite{chen2024comparative,serrano2025promptnorm,zamir2022restormer,gautam2025pureformer}, Multi-Dconv head Transposed Attention (MDTA) is well-suited for modeling global context and cross-channel correlations.

Conversely, for the high-frequency experts $E^{H}$, we utilize NAFNet~\cite{chen2022simple} blocks. As a lightweight CNN-based architecture, NAFBlock provides strong local inductive biases, making it highly suitable for removing sharp shadow boundaries while preserving fine textural details.

\begin{table*}
\centering
\setlength{\tabcolsep}{3pt}
\caption{Quantitative results on Ambient6K~\cite{vasluianu2024towards} (1280$\times$960 px. default size) and CL3AN~\cite{vasluianu2025after} (1920$\times$1440 px.) benchmark test sets, following the convention of Vasluianu \textit{et al.}~\cite{vasluianu2025after}. The \colorbox{Coral1}{best}, \colorbox{Goldenrod1}{second}, and \colorbox{LightGoldenrod1}{third best} results are highlighted. $^{\dagger}$ denotes MACs without additional models.}
\vspace{-2mm}
\resizebox{0.95\linewidth}{!}
{
\begin{tabular}{@{}r|c|c|c|c||ccc|ccc@{}}
\toprule
Method & \multicolumn{4}{c||}{General information} & \multicolumn{3}{c|}{Ambient6K \cite{vasluianu2024towards}} & \multicolumn{3}{c}{CL3AN \cite{vasluianu2025after}}\\
 name  &    Restoration Task  &  Type   & Prior  & MACs (G.) & \text{PSNR}$\uparrow$ & \text{SSIM}$\uparrow$ & \text{LPIPS}$\downarrow$ & \text{PSNR}$\uparrow$ & \text{SSIM}$\uparrow$ & \text{LPIPS}$\downarrow$\\
\midrule
unprocessed               & -   & -  & -                                        & -       & 13.403 & 0.652 & 0.250   & 10.837 & 0.447 & 0.518   \\
\midrule
NAFNet \cite{chen2022simple}   & \multirow{9}{*}{Multi-task IR}  & Conv. & RGB   & 15.92   & 20.580 & 0.808 & 0.142   & 19.476 & 0.709 & 0.249    \\
MPRNet \cite{zamir2021multi} &           &  Conv.          & RGB                & 37.21   & 20.947 & 0.820 & 0.129   & 18.453 & 0.688 & 0.291   \\
SFNet \cite{cui2023selective}  &         &  Transf.        & RGB + Freq.        & 31.27   & 20.519 & 0.812 & 0.141   & 18.382 & 0.686 & 0.291    \\
SwinIR \cite{liang2021swinir} &          &  Transf.        & RGB                & 37.81   & 20.528 & 0.817 & 0.131   & 16.386 & 0.643 & 0.372    \\
Uformer \cite{wang2022uformer} &         &  Transf.        & RGB                & 19.33   & 20.776 & 0.818 & 0.131   & 17.508 & 0.655 & 0.313    \\
Restormer \cite{zamir2022restormer}  &   &  Transf.        & RGB                & 35.31   & 21.141 & 0.817 & 0.132   & 18.560 & 0.691 & 0.278      \\
HINet \cite{chen2021hinet}  &            &  Conv.          & RGB                & 42.68   & 20.856 & 0.821 & 0.129   & 19.388 & 0.707 & 0.248      \\
MAMBAIR \cite{guo2024mambair} &          &  Transf. + SSM  & RGB                & 34.32   & -      & -     & -       & 18.970 & 0.704 & 0.254   \\
GRL \cite{li2023efficient} &             &  Transf.        & RGB                & 2.16    & -      & -     & -       & 18.089 & 0.672 & 0.308       \\
\midrule
Retinexformer \cite{cai2023retinexformer} & LLIE             & Transf. & RGB                & 4.86              & - & - & -  & 18.649 & 0.683 & 0.281    \\
\midrule
IFBlend \cite{vasluianu2024towards}   & \multirow{6}{*}{ALN} & Conv.  & RGB + Freq.         & 26.01             & 21.443 & 0.819 & 0.128 & 20.370 & 0.720 & 0.228    \\
PromptNorm \cite{serrano2025promptnorm}  &                   & Transf  & RGB + Geom.        & 13.49$^{\dagger}$ & \colorbox{LightGoldenrod1}{22.116} & 0.822 & 0.124 & -      & -     & -        \\
RLN$^2$-Sf \cite{vasluianu2025after} &                       & Conv.  & RGB + Freq.         & 7.89              & 21.333 & 0.819 & 0.128 & 20.128 & 0.730 & 0.223   \\
RLN$^2$-Lf \cite{vasluianu2025after} &                       & Conv.  & RGB + Freq.         & 22.72             & 21.712 & \colorbox{LightGoldenrod1}{0.825} & \colorbox{LightGoldenrod1}{0.120} & \colorbox{LightGoldenrod1}{20.523} & \colorbox{LightGoldenrod1}{0.746} & \colorbox{LightGoldenrod1}{0.208}    \\
DINOLight \cite{oh2026dinolight}  &                                                 & Transf.  & RGB + Freq. + DINO & 7.89$^{\dagger}$  &  \colorbox{Goldenrod1}{22.788} &  \colorbox{Goldenrod1}{0.838} &  \colorbox{Goldenrod1}{0.107} & \colorbox{Coral1}{21.107} & \colorbox{Goldenrod1}{0.764} & \colorbox{Goldenrod1}{0.188}    \\
\textbf{OmniLight (ours)}  &                                          & Transf.  & RGB + Freq. + DINO & 13.71$^{\dagger}$  & \colorbox{Coral1}{23.480} & \colorbox{Coral1}{0.848} & \colorbox{Coral1}{0.105} & \colorbox{Goldenrod1}{20.858} & \colorbox{Coral1}{0.769} & \colorbox{Coral1}{0.186}   \\
\bottomrule
\end{tabular}%
}
\label{table:aln_bench}
\end{table*}

\noindent\textbf{Routing heuristics.}
To generate the routing weights, our router $G(\cdot)$ relies on a composite global descriptor that can discriminate between degradations. In our implementation, the routing guidance vector $v$ is constructed by concatenating the Global Average Pooling (GAP) and Global Max Pooling (GMP) of both the final encoder feature $F_{enc}$ and the visual prior from the frozen DINOv2 encoder $F_{dino}$:

\begin{equation}
    \begin{aligned}
    v = \text{Concat}(\text{GAP}(F_{enc}), \text{GMP}(F_{enc}), \\
    \text{GAP}(F_{dino}), \text{GMP}(F_{dino})).
    \end{aligned}
\end{equation}

This routing design is adopted as a practical heuristic. The combination of average and max pooling aims to simultaneously capture the global statistical distribution of the scene (e.g., overall color temperature) and salient high-frequency anomalies (e.g., sharp shadow edges). Furthermore, incorporating the pre-extracted $F_{dino}$ provides a robust visual anchor, which helps stabilize the routing decisions across highly diverse lighting conditions. 

To optimize this routing mechanism and encourage stable feature modulation during training, we follow the overall expert training scheme and auxiliary load-balancing strategies proposed in MoCE-IR~\cite{zamfir2025complexity}.

\section{Experiments}
\label{sec:experiments}
\subsection{Datasets}
For the challenge, we use the NTIRE 2026 challenge training data for image shadow removal, ALN (white), and ALN (color), along with the publicly available WSRD+~\cite{vasluianu2023wsrd}, Ambient6K~\cite{vasluianu2024towards}, and CL3AN~\cite{vasluianu2025after} datasets to train DINOLight and OmniLight. For the benchmark results in \cref{subsec:aln}, we only use the public training datasets. The training data is cropped into 448×448-sized patches. We evaluate the methods using the NTIRE 2026 challenge test sets, and the WSRD+/Ambient6K/CL3AN test sets.

\begin{table}
\centering
\setlength{\tabcolsep}{5pt}
\caption{Quantitative results on the WSRD+~\cite{vasluianu2023wsrd} benchmark test set, following the convention of Xu \textit{et al.}~\cite{xu2025omnisr}. The \colorbox{Coral1}{best}, \colorbox{Goldenrod1}{second}, and \colorbox{LightGoldenrod1}{third best} results are highlighted.}
\vspace{-2mm}
\resizebox{0.65\linewidth}{!}
{
\begin{tabular}{@{}r|c|cc@{}}
\toprule
\multirow{2}{*}{Method} & \multirow{2}{*}{Year} & \multicolumn{2}{c}{WSRD+~\cite{vasluianu2023wsrd}}\\
\cmidrule{3-4}
 & & \text{PSNR}$\uparrow$ & \text{SSIM}$\uparrow$ \\
\midrule
DHAN~\cite{cun2020towards}                          & 2020 & 22.39 & 0.796 \\
Fu \textit{et al.}~\cite{fu2021auto}                         & 2021 & 21.66 & 0.752 \\
BMNet~\cite{zhu2022bijective}                       & 2022 & 24.75 & 0.816 \\
ShadowFormer~\cite{guo2023shadowformer}             & 2023 & 25.44 & 0.820 \\
ShadowRefiner~\cite{dong2024shadowrefiner}          & 2024 & 26.04 & 0.827 \\
OmniSR~\cite{xu2025omnisr}                          & 2025 & \colorbox{LightGoldenrod1}{26.07} & \colorbox{LightGoldenrod1}{0.835} \\
DINOLight~\cite{oh2026dinolight}                    & -    & \colorbox{Coral1}{27.17} & \colorbox{Coral1}{0.849} \\
\textbf{OmniLight (ours)}                           & -    & \colorbox{Goldenrod1}{26.90} & \colorbox{Goldenrod1}{0.848} \\
\bottomrule
\end{tabular}%
}
\label{table:wsrd_bench}
\end{table}

\begin{table}
\centering
\caption{NTIRE 2026 Challenge results across all three lighting-related tracks: Shadow Removal~\cite{ntire2026shadow}, ALN (White Lighting), and ALN (Color Lighting)~\cite{ntire2026ambient}. Results for the baseline \colorbox{PaleGreen1}{DINOLight} and our proposed \colorbox{LightBlue1}{OmniLight} are highlighted. [Keys: \textbf{\textcolor{red}{Best}}, \textbf{\textcolor{blue}{Second}}, \textbf{Third}, \underline{Fourth}.]}
\vspace{-2mm}
\label{table:ntire_all}
\resizebox{\linewidth}{!}{
\begin{tabular}{clccccc}
\toprule
\multicolumn{7}{c}{\textbf{Shadow Removal}} \\
\midrule
\textbf{Rank} & Team name & PSNR$\uparrow$ & SSIM$\uparrow$ & LPIPS$\downarrow$ & FID$\downarrow$ & \\
\midrule
\textbf{\textcolor{red}{1}} & AIT\_TUM & 26.680 & 0.874 & 0.058 & 26.135 & \\
\textbf{\textcolor{blue}{2}} & RAS & 26.137 & 0.866 & 0.071 & 30.466 & \\
\textbf{3} & \colorbox{PaleGreen1}{SNU-ISPL-B} & \colorbox{PaleGreen1}{25.943} & \colorbox{PaleGreen1}{0.867} & \colorbox{PaleGreen1}{0.085} & \colorbox{PaleGreen1}{28.053} & \\
\underline{4} & APRIL-AIGC & 26.449 & 0.848 & 0.079 & 29.650 & \\
5 & DiogenesCask & 26.432 & 0.868 & 0.086 & 33.640 & \\
6 & SNUCV & 25.300 & 0.864 & 0.068 & 29.995 & \\
7 & Shadow Breaker & 25.639 & 0.864 & 0.085 & 33.473 & \\
8 & CV\_SVNIT & 25.589 & 0.858 & 0.079 & 33.227 & \\
9 & \colorbox{LightBlue1}{SNU-ISPL-A} & \colorbox{LightBlue1}{25.818} & \colorbox{LightBlue1}{0.865} & \colorbox{LightBlue1}{0.088} & \colorbox{LightBlue1}{36.472} & \\
... & ... & ... & ... & ... & ... & \\
\midrule\midrule
\multicolumn{7}{c}{\textbf{ALN (White Lighting)}} \\
\midrule
\textbf{Percep. / Fidel.} & Team name & PSNR$\uparrow$ & SSIM$\uparrow$ & LPIPS$\downarrow$ & FID$\downarrow$ & MOS$\uparrow$ \\
\midrule
\textbf{\textcolor{red}{1}} / 6 & \colorbox{LightBlue1}{SNU-ISPL-A} & \colorbox{LightBlue1}{24.275} & \colorbox{LightBlue1}{0.854} & \colorbox{LightBlue1}{0.099} & \colorbox{LightBlue1}{54.169} & \colorbox{LightBlue1}{9.5} \\
\textbf{\textcolor{blue}{2}} / \textbf{\textcolor{red}{1}} & MILab\_ALN & 26.680 & 0.873 & 0.083 & 50.218 & 9 \\
\textbf{3} / \textbf{\textcolor{blue}{2}} & ACVLAB & 24.875 & 0.864 & 0.089 & 49.838 & 8.75 \\
\underline{4} / 11 & MiPorAlgo & 23.141 & 0.763 & 0.169 & 59.441 & 8.5 \\
5 / \textbf{3} & OUT\_OF\_MEMORY & 25.416 & 0.860 & 0.094 & 54.496 & 8.25 \\
6 / 5 & \colorbox{PaleGreen1}{SNU-ISPL-B} & \colorbox{PaleGreen1}{24.300} & \colorbox{PaleGreen1}{0.856} & \colorbox{PaleGreen1}{0.101} & \colorbox{PaleGreen1}{51.883} & \colorbox{PaleGreen1}{8} \\
... & ... & ... & ... & ... & ... & ... \\
\midrule\midrule
\multicolumn{7}{c}{\textbf{ALN (Color Lighting)}} \\
\midrule
\textbf{Percep. / Fidel.} & Team name & PSNR$\uparrow$ & SSIM$\uparrow$ & LPIPS$\downarrow$ & FID$\downarrow$ & MOS$\uparrow$ \\
\midrule
\textbf{\textcolor{red}{1}} / \underline{4} & MiPorAlgo & 21.305 & 0.629 & 0.248 & 92.900 & 8.25 \\
\textbf{\textcolor{blue}{2}} / \textbf{\textcolor{red}{1}} & \colorbox{PaleGreen1}{SNU-ISPL-B} & \colorbox{PaleGreen1}{22.038} & \colorbox{PaleGreen1}{0.753} & \colorbox{PaleGreen1}{0.186} & \colorbox{PaleGreen1}{82.013} & \colorbox{PaleGreen1}{7.75} \\
\textbf{3} / \textbf{3} & ACVLAB & 21.432 & 0.743 & 0.196 & 88.200 & 7.5 \\
\underline{4} / \textbf{\textcolor{blue}{2}} & \colorbox{LightBlue1}{SNU-ISPL-A} & \colorbox{LightBlue1}{21.900} & \colorbox{LightBlue1}{0.744} & \colorbox{LightBlue1}{0.204} & \colorbox{LightBlue1}{86.654} & \colorbox{LightBlue1}{7.5} \\
... & ... & ... & ... & ... & ... & \\
\bottomrule
\end{tabular}
}
\end{table}

\subsection{Implementation Details}
We implement OmniLight using PyTorch and train with 2 NVIDIA GeForce RTX 3090 GPUs. The training patches are augmented with random flipping or rotations.

\noindent\textbf{Loss function.} We optimize OmniLight using a combination of L1 loss, Multi-Scale Structural Similarity (MS-SSIM) loss~\cite{wang2003multiscale}, and an auxiliary load-balancing loss~\cite{zamfir2025complexity}. The total objective function is defined as:
\begin{equation}
    \mathcal{L}_{total} = \mathcal{L}_{L1} + \alpha \cdot\mathcal{L}_{MS-SSIM} + \beta \cdot \mathcal{L}_{Aux},
\end{equation}
The hyper-parameters $\alpha$ and $\beta$ are weighting coefficients to balance the structural and routing objectives, respectively. We set $\alpha$ as 0.25 and $\beta$ as 0.01 in this work. 

\subsubsection{Training Strategy}
Due to the divergent characteristics of the target datasets, which range from hard shadow boundaries to complex multi-color entanglements, we observed that joint training from scratch poses a potential risk of optimization instability. To proactively stabilize the training process, we adopt a two-stage initialization and fine-tuning strategy. 

\noindent\textbf{1st stage:} We initialize the network by training exclusively on the Ambient6K dataset. We use the AdamW~\cite{loshchilov2017decoupled} optimizer ($\beta_1=0.9$, $\beta_2=0.999$) and an initial learning rate of 2e-4 that is gradually decayed to 1e-6 using cosine annealing~\cite{loshchilov2016sgdr}, with a batch size of 2 for 100 epochs. During this stage, the WD-MoE branch is kept frozen. 
By excluding complex chromatic shifts in this initial phase, the model first learns to resolve structural shadow and lighting intensity variations, providing a stable baseline.

\noindent\textbf{2nd stage:} Subsequently, using these pre-trained weights, we jointly fine-tune the network on the combined dataset comprising NTIRE 2026 challenge~\cite{ntire2026ambient,ntire2026shadow} training datasets, WSRD+~\cite{vasluianu2023wsrd}, Ambient6K~\cite{vasluianu2024towards}, and CL3AN~\cite{vasluianu2025after}, with a batch size of 8 for 100 epochs. 

Since our WD-MoE branch utilizes three experts per sub-band, we observed that a small batch size of 2 is often insufficient to distribute samples across all experts in a single forward pass, which increases the risk of unstable gradients and optimization difficulties. Therefore, a larger batch size helps ensure that each expert receives adequate gradients, facilitating the effective operation of the auxiliary load-balancing loss. While hardware limitations restricted our maximum batch size to 8, a larger batch size is generally expected to further enhance the effectiveness of the load-balancing mechanism.

In this phase, we apply a linear learning rate warm-up from 1e-6 to 1e-4 over the initial 10 epochs, followed by cosine annealing decay to 1e-6 for the remaining 90 epochs. This warm-up period stabilizes the early optimization process, as it protects the pre-trained weights from being disrupted by sudden exposure to complex target domains.
Simultaneously, it allows the WD-MoE branches to achieve stable convergence.

For a fair comparison between both methods, we follow the training strategy of OmniLight for DINOLight, with two necessary modifications. First, we exclude the auxiliary load-balancing loss term $\mathcal{L}_{Aux}$, as it is specific to the MoE routing. Second, during the entire training process, DINOLight is trained exclusively on the dataset corresponding to its specific target task, rather than the combined dataset.

\subsubsection{Inference}
At inference, we adopt a sliding window inference strategy to split the high-resolution
input into fixed-size patches identical to the training resolution and process them. The user is able to decide the overlap ratio, which has a trade-off between performance metrics and inference speed. Runtime varies from 1.6s to 3.1s and 1.6s to 2.6s on an RTX 3090 for input resolution of 720$\times$960 and 768$\times$1024, respectively, according to the overlap ratio.

\begin{figure*}[t]
\centering
   \includegraphics[width=0.95\textwidth]{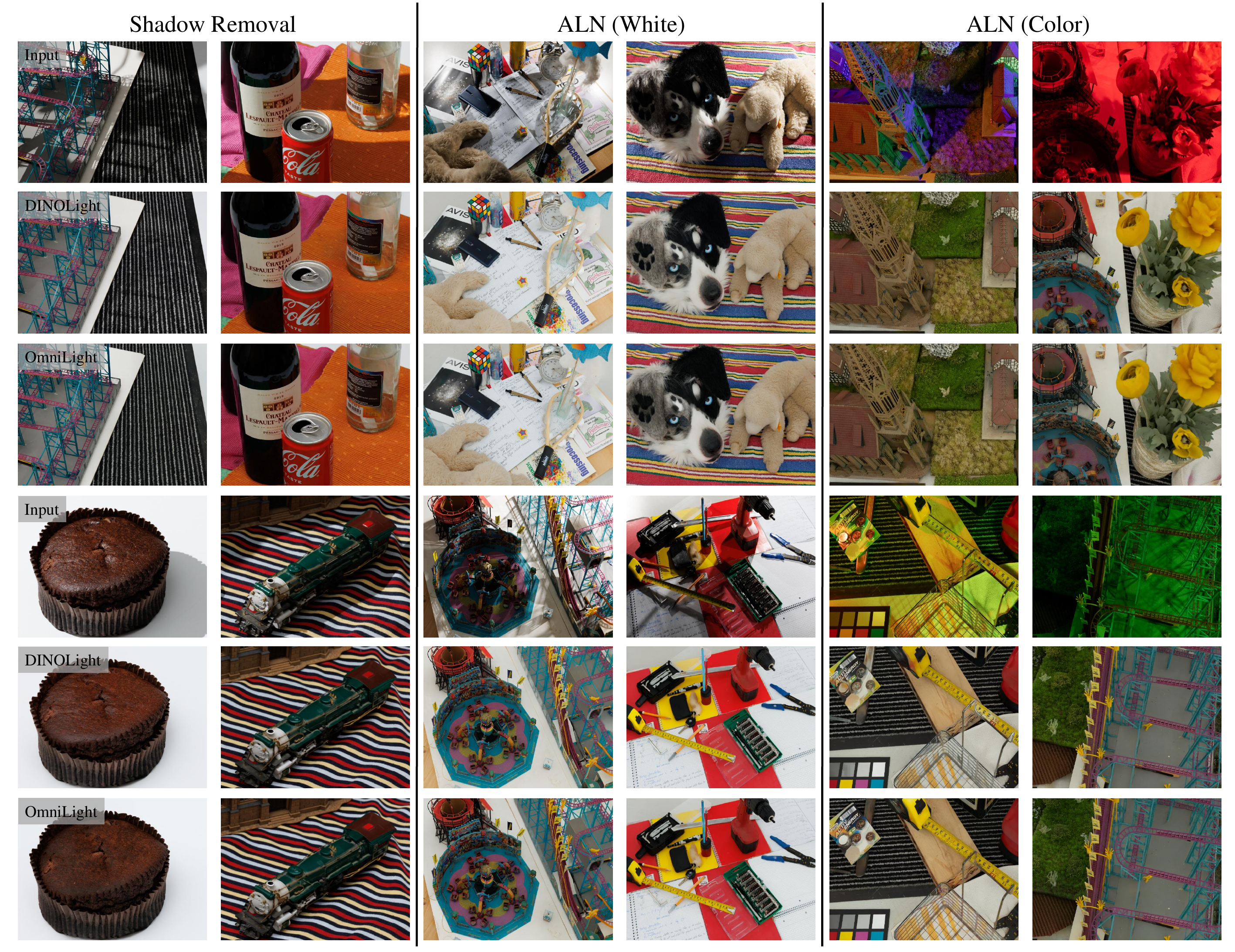}
   \hfil
\vspace{-3mm}
\caption{Qualitative results of lighting-related image restoration of NTIRE 2026 Challenge~\cite{ntire2026ambient,ntire2026shadow} test sets.}
\label{figure:comparison}
\end{figure*}

\subsection{Results}

\noindent\textbf{Evaluation metrics.} For quantitative evaluation, we use three standard metrics: Peak Signal-to-Noise Ratio (PSNR) and Structural Similarity Index Measure (SSIM) for pixel-level and structural fidelity, and Learned Perceptual Image Patch Similarity (LPIPS)~\cite{zhang2018unreasonable} to assess perceptual quality. The challenge~\cite{ntire2026ambient,ntire2026shadow} also utilizes Frechet Inception Distance (FID)~\cite{heusel2017gans} and Mean Opinion Score (MOS) for further evaluation.

\subsubsection{ALN \& Shadow Removal Quantitative Results}
\label{subsec:aln}
Following previous benchmarking literature by Vasluianu \textit{et al.}~\cite{vasluianu2025after}, we show comparative results of OmniLight on Ambient6K~\cite{vasluianu2024towards} and CL3AN~\cite{vasluianu2025after} against existing methods and baselines~\cite{vasluianu2024towards,vasluianu2025after,chen2022simple,zamir2022restormer,chen2021hinet,zamir2021multi,cui2023selective,liang2021swinir,wang2022uformer,guo2024mambair,li2023efficient,cai2023retinexformer,serrano2025promptnorm,oh2026dinolight}. 
As shown in \cref{table:aln_bench}, both the baseline DINOLight~\cite{oh2026dinolight} and the proposed OmniLight demonstrate state-of-the-art performance, significantly outperforming all prior methods across Ambient6K and CL3AN benchmarks. While the task-specific baseline naturally achieves the highest PSNR on CL3AN (+0.25~dB) due to its direct optimization for a single target distribution, our unified OmniLight framework secures the top performance on Ambient6K (+0.69~dB in PSNR), and also surpasses DINOLight in structural and perceptual metrics (SSIM, LPIPS) even on CL3AN.

We also evaluate our method on WSRD+~\cite{vasluianu2023wsrd} and compare it against \cite{cun2020towards,fu2021auto,zhu2022bijective,guo2023shadowformer,dong2024shadowrefiner,xu2025omnisr,oh2026dinolight}, following Xu \textit{et al.}~\cite{xu2025omnisr}. \cref{table:wsrd_bench} shows that OmniLight achieves competitive performance compared to the specialized baseline, which achieves the best performance among all methods.

These results demonstrate that although OmniLight is designed to generalize across multiple, highly diverse lighting conditions as a single model, it successfully achieves highly competitive or even superior quantitative metrics compared to methods that are explicitly tailored for a specific task.

\subsubsection{NTIRE 2026 Challenge}
We also participated in the NTIRE 2026 Shadow Removal, ALN (White Lighting), and ALN (Color Lighting) challenge with the baseline, and our proposed OmniLight. 

\noindent\textbf{Quantitative results.}
The quantitative results of the challenge are presented in \cref{table:ntire_all}. Our proposed OmniLight achieved \textbf{\textcolor{red}{1st (Perceptual)}} place in ALN White Lighting, alongside \textbf{\textcolor{blue}{2nd (Fidelity)}} and \underline{4th (Perceptual)} places in ALN Color Lighting. Furthermore, the baseline DINOLight secured \textbf{\textcolor{red}{1st (Fidelity)}} and \textbf{\textcolor{blue}{2nd (Perceptual)}} places in ALN Color Lighting, as well as \textbf{3rd Overall} in Shadow Removal. These consistent top-tier rankings demonstrate the robust performance and perceptual quality of both the baseline and our unified approach. Detailed results are available in the official challenge reports~\cite{ntire2026ambient,ntire2026shadow}.

\noindent\textbf{Qualitative results.}
\cref{figure:comparison} provides a qualitative comparison of the proposed OmniLight and the baseline across the three lighting-related restoration tasks. As illustrated in the visual examples, OmniLight consistently generates highly pleasing and natural-looking results across all diverse scenarios, comparable to or sometimes even better than the specialized model DINOLight. It effectively removes hard shadows and neutralizes complex color casts while faithfully preserving the underlying image details and original textures, confirming its strong generalization capability and robustness in handling various unconstrained illumination conditions.

\subsection{Analysis on Specialized and Unified Learning}

By comparing the metrics of the baseline~\cite{oh2026dinolight} and our proposed method in \cref{table:aln_bench,table:wsrd_bench,table:ntire_all}, we observe a typical trade-off behavior between task-specific precision and cross-domain generalization. Consequently, this comparison effectively functions as an ablation study for our unified framework. Visual comparisons are available in \cref{figure:comparison}.

\begin{figure}[t]
\centering
   \includegraphics[width=0.45\textwidth]{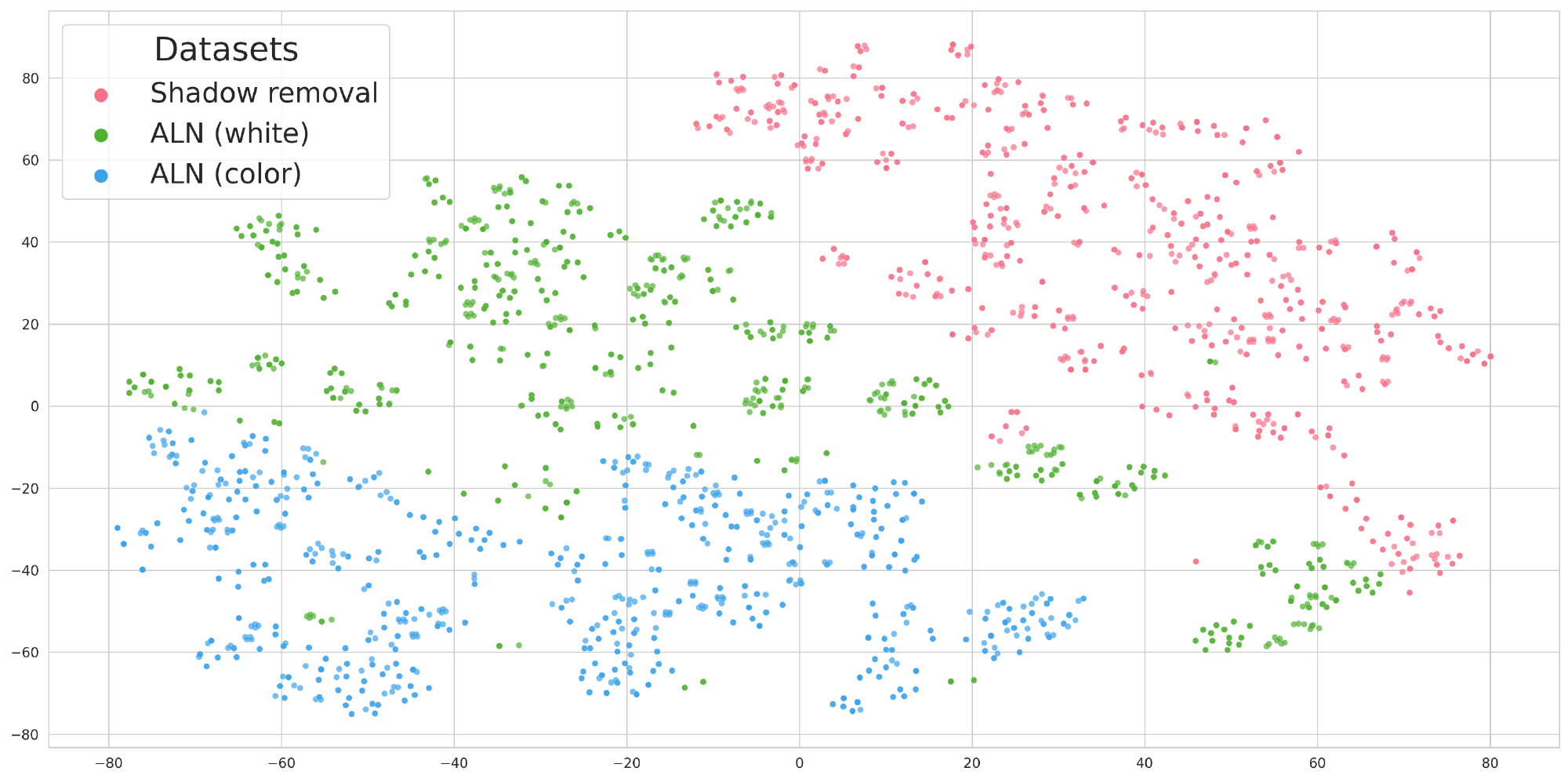}
   \hfil
\vspace{-2mm}
\caption{t-SNE visualization~\cite{van2008visualizing} of the routing guidance vector.}
\label{figure:tsne}
\end{figure}

\begin{figure}[t]
\centering
   \includegraphics[width=0.45\textwidth]{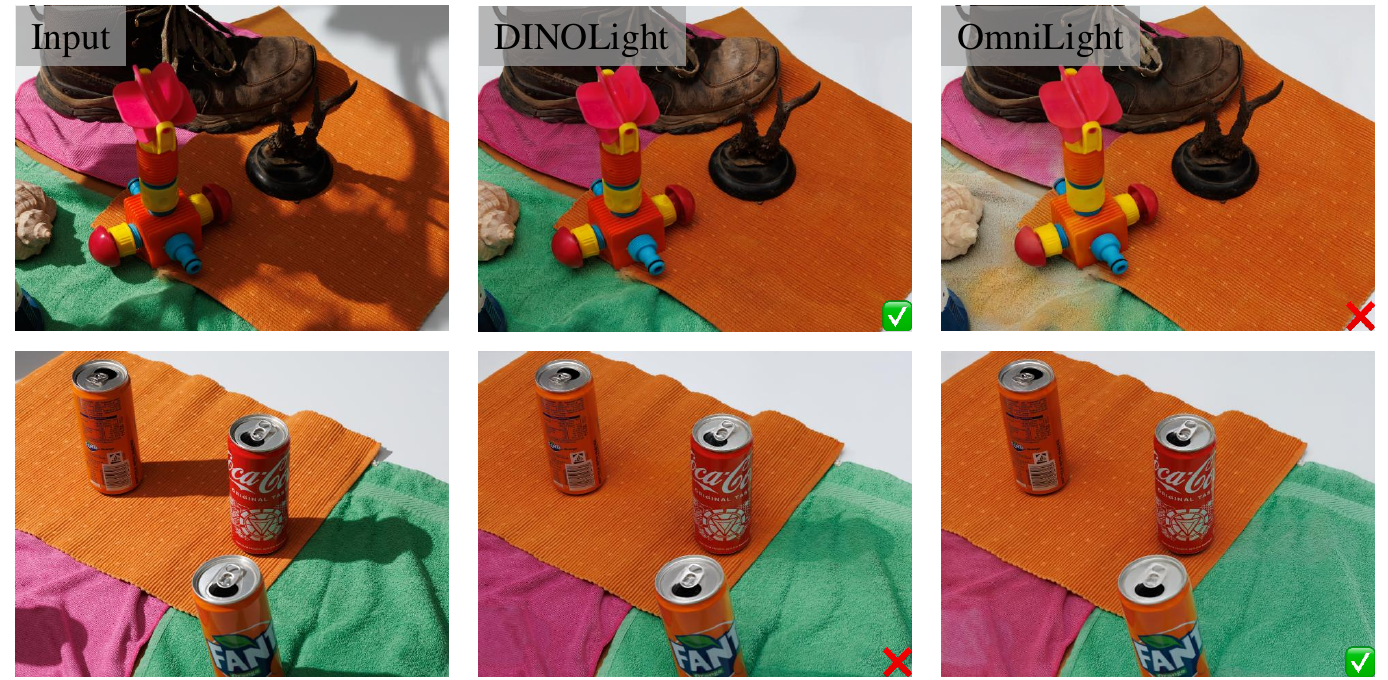}
   \hfil
\vspace{-2mm}
\caption{Failure and success cases of restoration from NTIRE 2026 Image Shadow removal test set for OmniLight.}
\label{figure:failure}
\end{figure}

\subsubsection{Specialized Model} 
By solely optimizing for the unique shadow characteristics and illumination distributions of individual target domain and datasets, DINOLight generally achieves better quantitative results than OmniLight. However, this high degree of specialization implies that DINOLight can occasionally be more sensitive to out-of-distribution variations or edge cases within unconstrained environments. Furthermore, DINOLight inherently exhibits limited adaptability to untrained tasks, causing substantial performance degradation and failing to ensure consistent visual quality in such cases.

\subsubsection{Unified Model} 
Conversely, our unified OmniLight model demonstrates remarkable robustness across a diverse spectrum of complex lighting conditions. By jointly learning from the combined datasets, OmniLight’s WD-MoE architecture acquires a comprehensive illumination prior encompassing both structural restoration and chromatic adaptation. This extensive data exposure acts as a strong regularizer; while it yields a marginal drop in some dataset-specific metrics compared to DINOLight, it effectively ensures a more consistent performance across varying illumination scenarios.

\noindent\textbf{Analysis on routing guidance vectors.} 
To validate whether the proposed WD-MoE effectively handles increasingly complex lighting conditions, we investigate the behavior of the routing guidance vector $v$ of OmniLight across the diverse target datasets.

We extract the vectors from the bottleneck WD-MoE module of our challenge model evaluated on the challenge datasets and visualize their distribution using t-SNE~\cite{van2008visualizing}, as shown in \cref{figure:tsne}. The plot reveals explicitly distinct and well-separated clusters depending on the datasets, naturally transitioning from simple shadows to multi-source color casts. This indicates that the routers are provided with degradation-aware guidance signals that successfully recognize the global context and the specific type of illumination degradation.

\subsection{Limitations and Future Work}
Despite the robust generalization capabilities of OmniLight, which yields highly competitive and occasionally superior results compared to DINOLight, we observe a specific failure case related to task interference. Examples of both failure and success cases are provided in \cref{figure:failure}. In shadow removal, which frequently features simple object geometries illuminated by a single light source, OmniLight sometimes introduces unintended color shifts in localized regions even when only a simple luminance correction is required. This results in a relatively larger drop in performance and perceived visual quality compared to the two other ALN tasks. 

We suspect that this artifact stems from an imperfect decoupling of the extreme chromatic adaptations learned from the colored ALN dataset. 
Achieving perfect disentanglement between highly divergent lighting domains without sacrificing unified representation remains a challenging problem. Furthermore, devising novel yet sophisticated routing strategies is necessary to resolve expert load imbalance and efficiently maximize the dynamic capacity of the network.
We leave these as an important direction for future research.
\section{Conclusion}
\label{sec:conclusion}
In this paper, we unified shadow removal and ALN under a comprehensive lighting-related image restoration problem. To tackle the diverse lighting complexities within this domain, we investigated two distinct approaches: an existing specialized baseline, DINOLight, and our proposed generalized all-in-one architecture, OmniLight. By leveraging WD-MoE, OmniLight effectively mitigates negative transfer across diverse lighting tasks by learning a universal representation guided by domain-aware signals. Our evaluations, highlighted by outstanding performance in lighting-related NTIRE 2026 image restoration challenges, reveal a clear trade-off: while specialized models maximize quantitative fidelity on specific data distributions, our unified OmniLight framework delivers a consistent and generalized solution for real-world lighting scenarios while maintaining competitive performance. Despite its broad generalization, OmniLight achieves state-of-the-art and near-state-of-the-art results on standard benchmarks such as WSRD+, Ambient6K, and CL3AN. We hope this work serves as a cornerstone for versatile, all-in-one image restoration models robust to unconstrained real-world lighting.

\noindent{\textbf{Acknowledgments.}} This work was supported by Samsung Electronics Co., Ltd(IO251211-14315-01), and in part by the BK21 FOUR program of the Education and Research Program for Future ICT Pioneers, Seoul National University in 2026.

{
    \small
    \bibliographystyle{ieeenat_fullname}

}


\end{document}